\documentclass[10pt,twocolumn,letterpaper]{article}

\usepackage[pagenumbers]{cvpr} 

\usepackage{graphicx}
\usepackage{amsmath}
\usepackage{amssymb}
\usepackage{booktabs}

\usepackage[pagebackref,breaklinks,colorlinks]{hyperref}

\usepackage[capitalize]{cleveref}
\crefname{section}{Sec.}{Secs.}
\Crefname{section}{Section}{Sections}
\Crefname{table}{Table}{Tables}
\crefname{table}{Tab.}{Tabs.}

\def\cvprPaperID{11857} 
\def\confName{CVPR}
\def\confYear{2022}

\usepackage{overpic}
\usepackage{enumitem} 
\usepackage{overpic} 
\usepackage{color}

\definecolor{turquoise}{cmyk}{0.65,0,0.1,0.3}
\definecolor{purple}{rgb}{0.65,0,0.65}
\definecolor{dark_green}{rgb}{0, 0.5, 0}
\definecolor{orange}{rgb}{0.8, 0.6, 0.2}
\definecolor{red}{rgb}{0.8, 0.2, 0.2}
\definecolor{darkred}{rgb}{0.6, 0.1, 0.05}
\definecolor{blueish}{rgb}{0.0, 0.3, .6}
\definecolor{light_gray}{rgb}{0.7, 0.7, .7}
\definecolor{pink}{rgb}{1, 0, 1}
\definecolor{greyblue}{rgb}{0.25, 0.25, 1}

\usepackage{blindtext}

\renewcommand{\paragraph}[1]{\vspace{1em}\noindent\textbf{#1}.}
\begin{document}
\title{Segmentation in Style: Unsupervised Semantic Image Segmentation with Stylegan and CLIP}
\author{Daniil Pakhomov\\
Johns Hopkins University\\
{\tt\small dpakhom1@jhu.edu}
\and
Sanchit Hira\\
Johns Hopkins University\\
{\tt\small shira2@jh.edu}
\and
Narayani Wagle\\
Johns Hopkins University\\
{\tt\small nwagle@jhu.edu}
\and
Kemar E. Green\\
Johns Hopkins University\\
{\tt\small kgreen66@jhmi.edu}
\and
Nassir Navab\\
Johns Hopkins University\\
{\tt\small nassir.navab@jhu.edu}
}
\maketitle


\begin{abstract}
   We introduce a method that allows to automatically segment images into semantically meaningful regions without human supervision. The derived regions are consistent across different images and coincide with human-defined semantic classes on some datasets. The method is particularly useful in cases where the labelling and definition of semantic regions pose a challenge for humans. In our work, we use pretrained StyleGAN2~\cite{karras2020analyzing} generative model: clustering in the feature space of the generative model allows to discover semantic classes. Once classes are discovered, a synthetic dataset with generated images and corresponding segmentation masks is created. After that a segmentation model is trained on the synthetic dataset and is able to generalize to real images. Additionally, by using CLIP~\cite{radford2021learning} we are able to use prompts defined in a natural language to discover some desired semantic classes. We test our method on publicly available datasets and show state-of-the-art results. The source code for the experiments
reported in the paper has been made public~\footnote{\url{https://github.com/warmspringwinds/segmentation_in_style}}
   
\end{abstract}

\section{Introduction}

\begin{figure}[t]
\begin{center}
\includegraphics[width=1.0\linewidth]{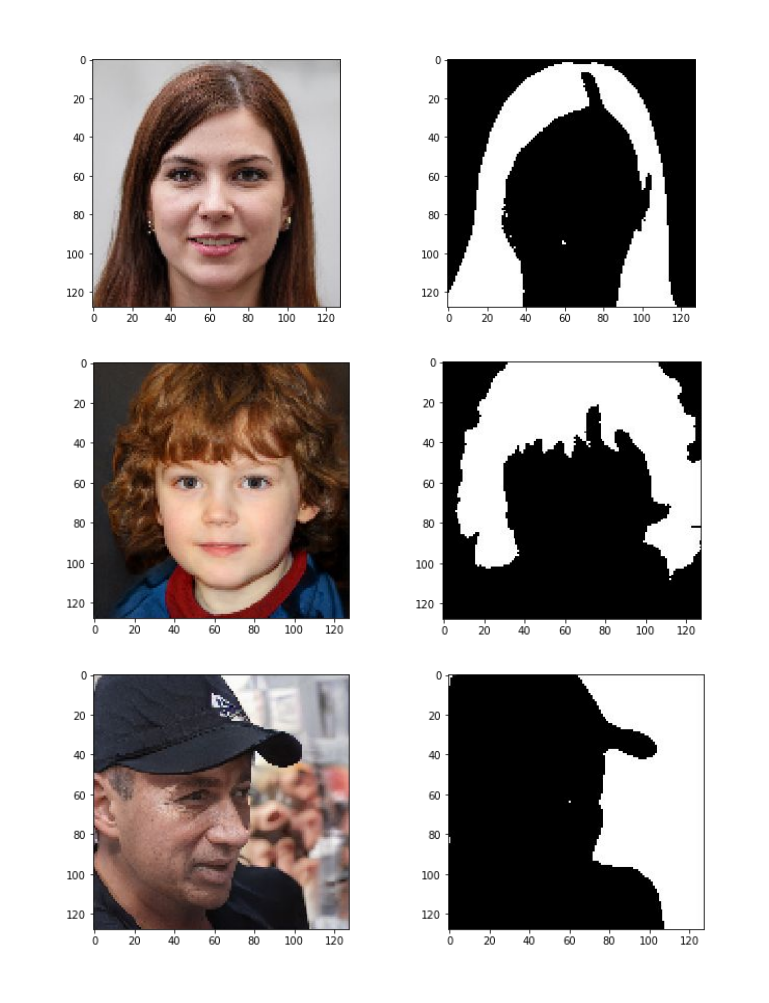}
\end{center}
   \caption{Example of synthetic image and annotation pairs created with our method for hair segmentation (first and second row) and background segmentation (third row).}
\label{fig:hair}
\label{fig:onecol}
\end{figure}

\begin{figure*}
\begin{center}
\includegraphics[width=.9\linewidth]{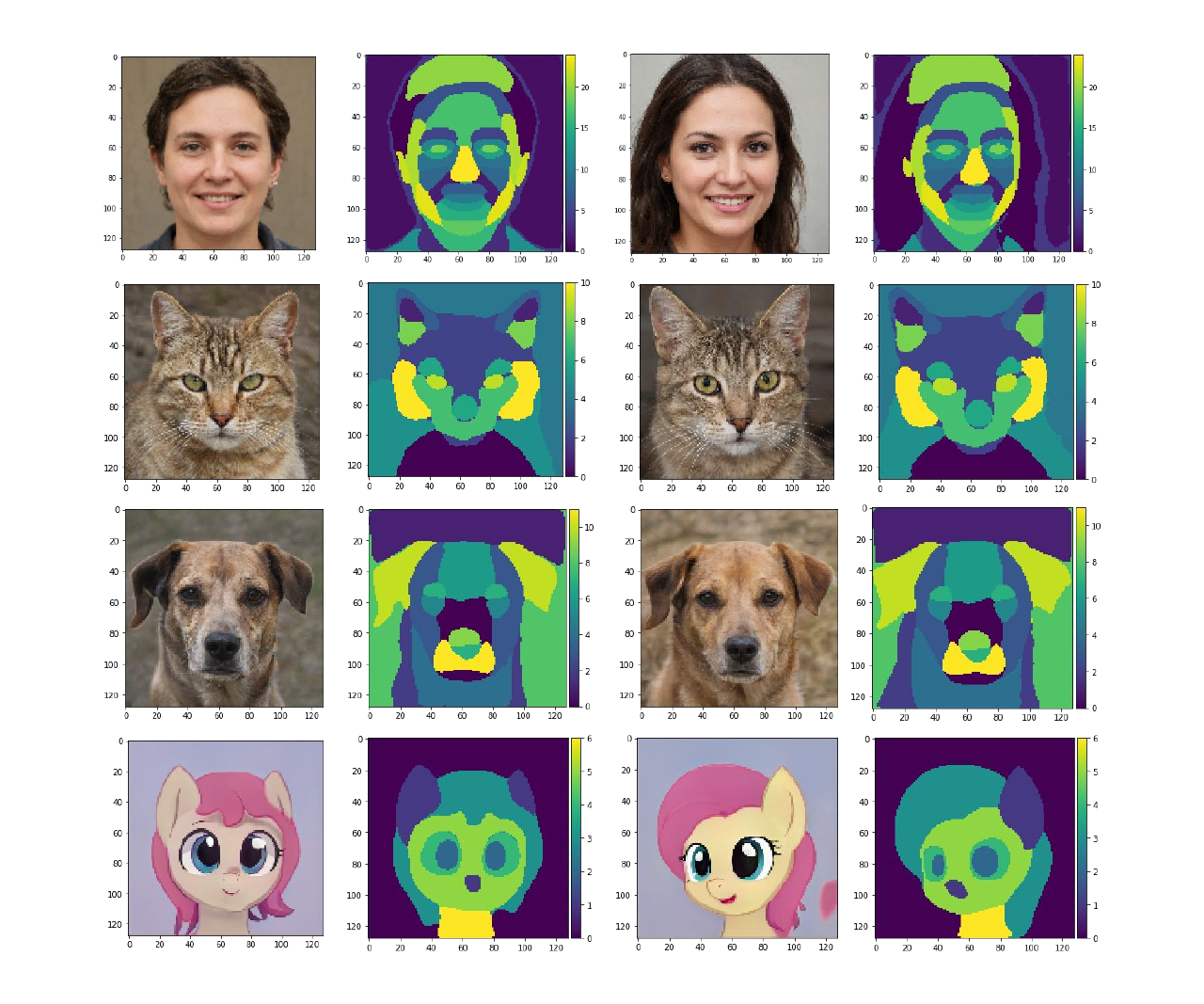}
\end{center}
   \caption{Figure demonstrates synthetic dataset examples   generated using our method with a stylegan model pretrained on Flickr-Faces-HQ (FFHQ)~\cite{karras2018style} dataset (first row), Animal faces (AFHQ)~\cite{choi2020stargan} dataset (second and third rows for cats and dogs) and a cartoon dataset~\cite{pony}. Semantic regions proposed by the network are consistent across samples even though there is no clear visual border between most semantic classes: it is usually hard for human annotators to consistently label examples like this while our method works well.}
\label{fig:parts}
\end{figure*}

The development of deep convolutional neural networks ~\cite{he2016deep,xie2017aggregated,hu2018squeeze} has fueled remarkable progress in semantic segmentation and pushed state-of-the-art on a variety of datasets~\cite{chen2018deeplab,zhao2017pyramid,yu2017dilated}. While the introduced methods show impressive results, their applicability is limited by a very slow process of image annotation~\cite{cordts2016cityscapes}. This issue is more pronounced in the medical imaging community where a domain knowledge is required to perform annotations. Consistent labeling can be a difficult task for a team of annotators working on the same dataset as most semantic classes lack clear visual boundaries. This can lead to some heterogeneity in the labels which can affect the overall model's performance. At the same time most deep learning approaches require a considerable amount of training data to reach their best performance~\cite{he2016deep}. In this paper, we introduce a method that does not require any labeling, allows to discover consistent semantic regions across images, outperforms semi-supervised methods that require a small number of annotations and is competitive with fully supervised image segmentation methods.

\begin{figure*}
\begin{center}
\includegraphics[width=.9\linewidth]{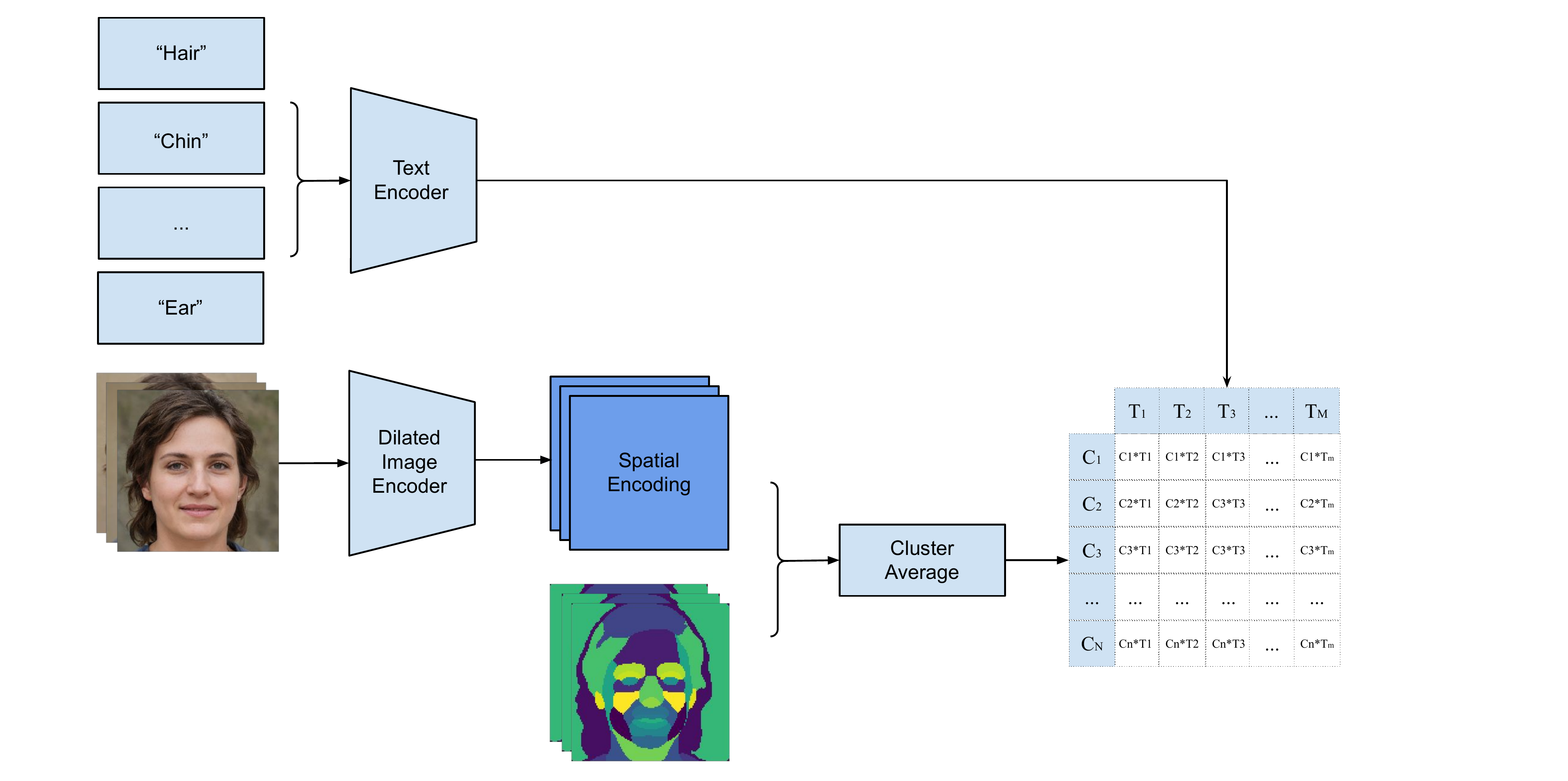}
\end{center}
   \caption{Figure demonstrates a process of classification of previously discovered clusters. Given a set of text prompts of desired classes and a set of generated images with corresponding clusters, we embed both text and image regions using pretrained text encoder and image encoders of CLIP~\cite{radford2021learning}. After that we compute pairwise dot products between text and cluster embeddings. Each cluster is assigned to a text prompt that results in a biggest dot product value. For example, in a given set of images all clusters containing hair will be classified as ``hair''.}
\label{fig:dilated_clip}
\end{figure*}

\begin{figure*}
\begin{center}
\includegraphics[width=.9\linewidth]{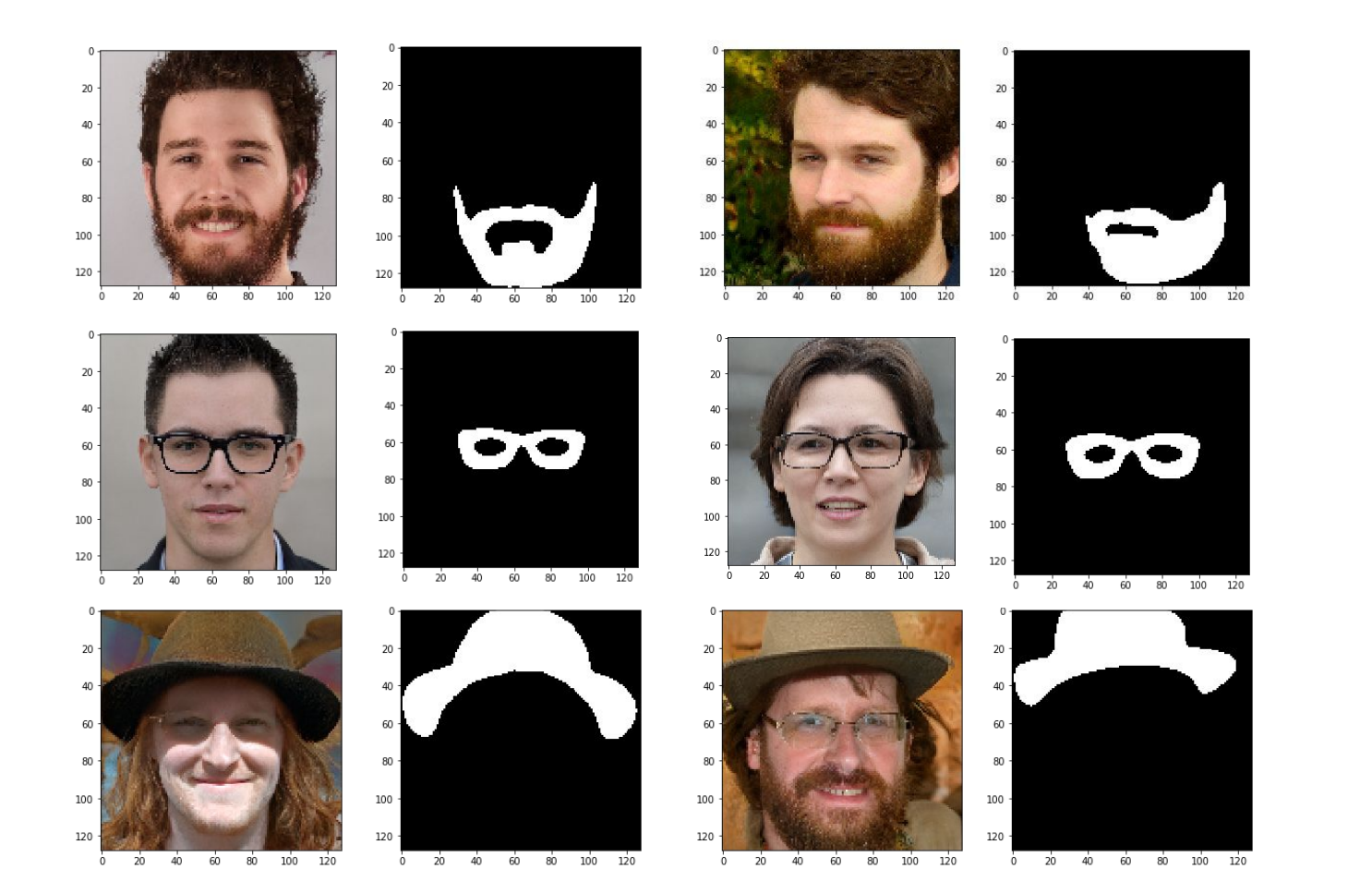}
\end{center}
   \caption{Example of synthetic image and annotation pairs created for rare classes with our method with CLIP~\cite{radford2021learning} using text prompts ``a person with a beard'' (first row), ``a person with glasses'' (second row), ``a person wearing a hat'' (third row). Interestingly, the discovered class representing glasses does not include eye regions which is consistent with human annotations of glasses for some datasets like CelebAMask-HQ~\cite{CelebAMask-HQ}.}
\label{fig:clip_results}
\end{figure*}

In our work we compare our method against semi-supervised learning approaches (SSL) which use a small number of annotations. This serves as a very strong baseline for our approach and we show that we outperform recent state-of-the-art methods~\cite{li2021semantic, zhang2021datasetgan} even though we do not use any annotations. Our method allows to suggest consistent semantic classes and we show that on some datasets they coincide with annotations provided by human which allows us to measure its accuracy. As it can be seen our method allows to discover high quality semantic regions that coincide with human defined regions (see Fig.~\ref{fig:hair}) and also suggest consistent semantic regions that might be hard for human to discover and label (see Fig.~\ref{fig:parts}). Additionally, we introduce a method that allows us to use natural language to discover certain rare semantic regions like beard, glasses and hats (see Fig.~\ref{fig:clip_results}). Finally, we show how we can automatically classify each of the discovered semantic regions and assign a prompt defined in a natural language to each of them (see Fig.~\ref{fig:dilated_clip}). To demonstrate applicability of our method across other domains, we carry out experiments on an eye segmentation dataset~\cite{palmero2020openeds2020} and show that our discovered semantic regions replicate human annotations  (see Fig.~\ref{fig:openeds}) and that the accuracy of our provided segmentation masks is competitive with fully supervised methods.


\begin{figure*}
\begin{center}
\includegraphics[width=.9\linewidth]{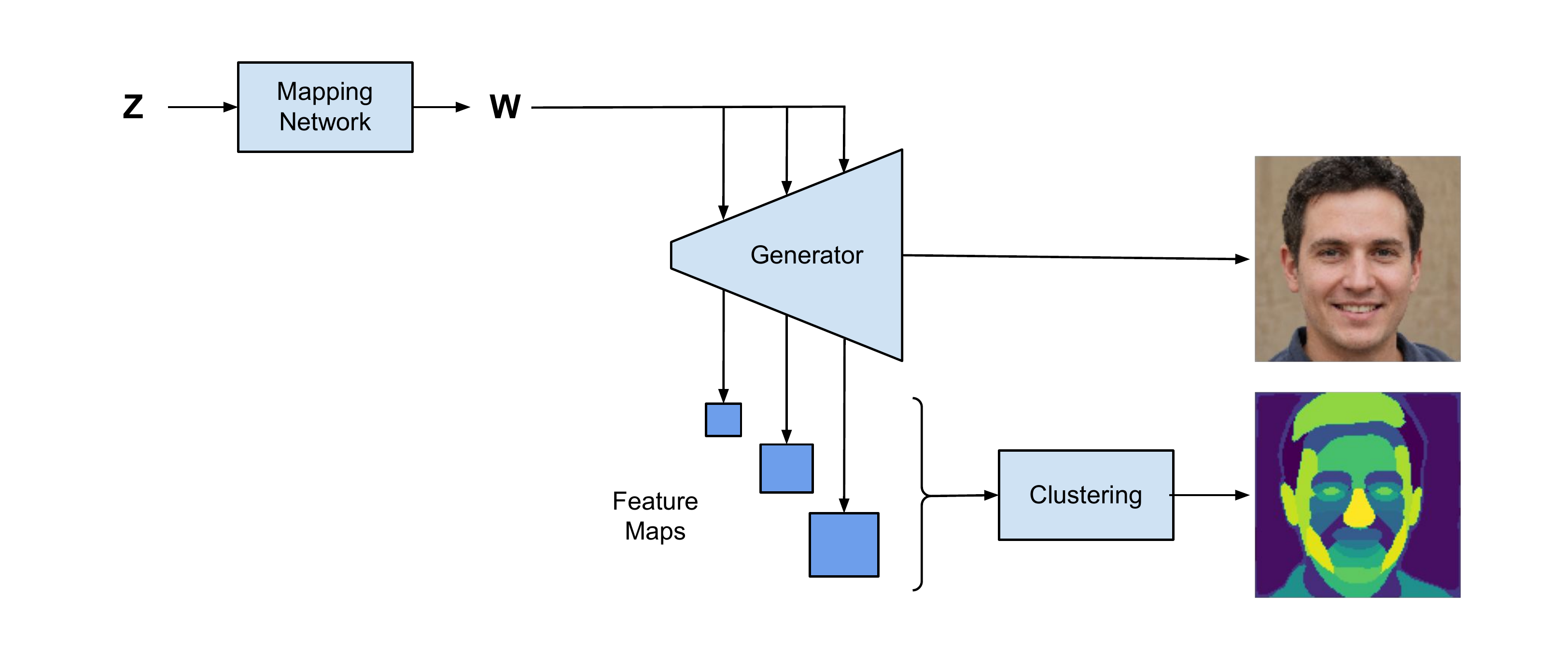}
\end{center}
   \caption{Figure demonstrates a process of creating of a synthetic dataset with semantic annotations for our approach. Stylegan consists of mapping network and generator networks which allow to create sythetic images. First we generate $N$ images and save their intermediate feature maps produced by the generator network. Clustering allows us to find semantic regions of the generated images. After the clusters are found, a much bigger set of images is generated and using their feature maps we attribute each pixel to one of the previously discovered semantic clusters. A segmentation network is later on trained on the synthetic dataset. }
\label{fig:stylegan_no_manipulation}
\end{figure*}

\begin{figure*}
\begin{center}
\includegraphics[width=.9\linewidth]{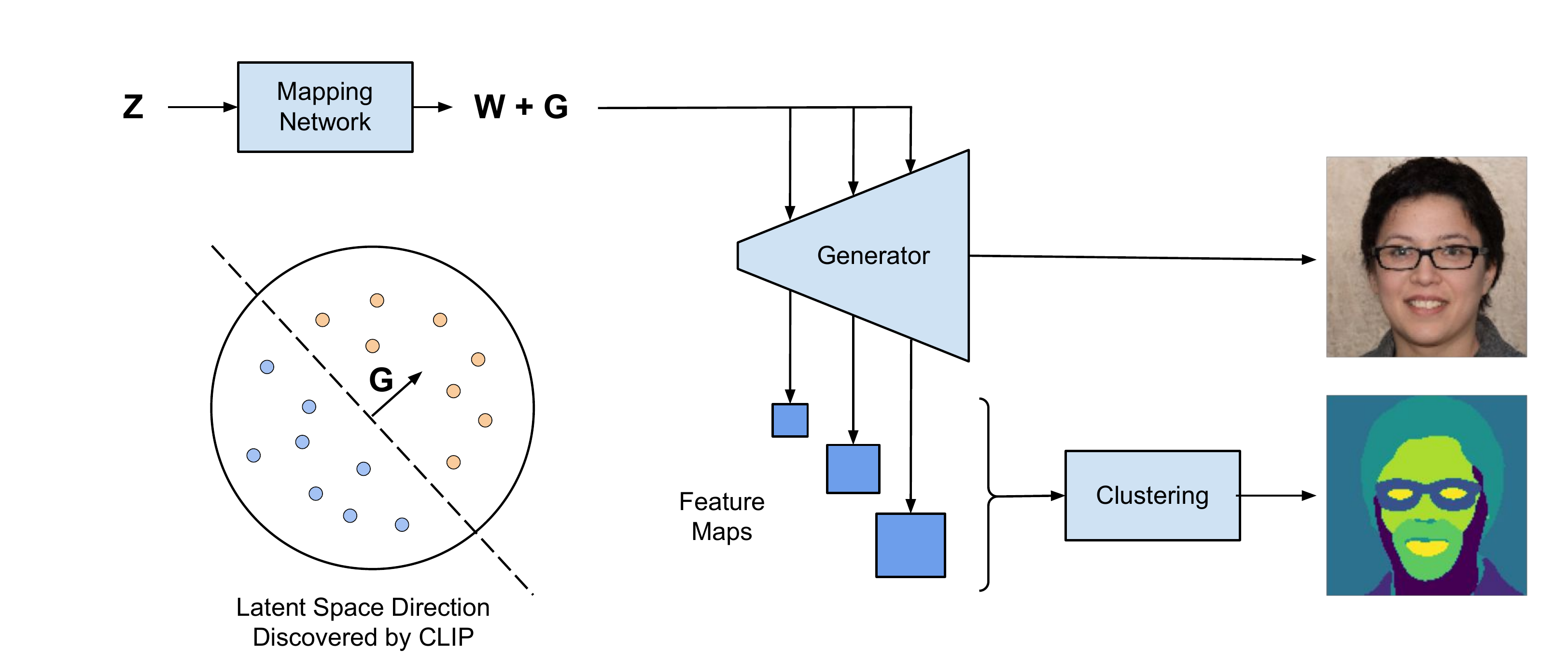}
\end{center}
   \caption{Figure demonstrates a process of creating a synthetic annotations for rare classes that were not discovered during clustering. First, a latent direction in the Stylegan is learnt that adds a desired semantic class to almost every generated sample using the method of~\cite{shen2020interfacegan, patashnik2021styleclip}. In this case, the vector $G$ represents the a text promt ``a person with glasses''. After that, almost every sample has the desired attribute and it naturally appears as one of the clusters. As it can be seen, semantic region representing glasses is indeed represented by one of the clusters.}
\label{fig:method_latent}
\end{figure*}

\begin{figure*}
\begin{center}
\includegraphics[width=.9\linewidth]{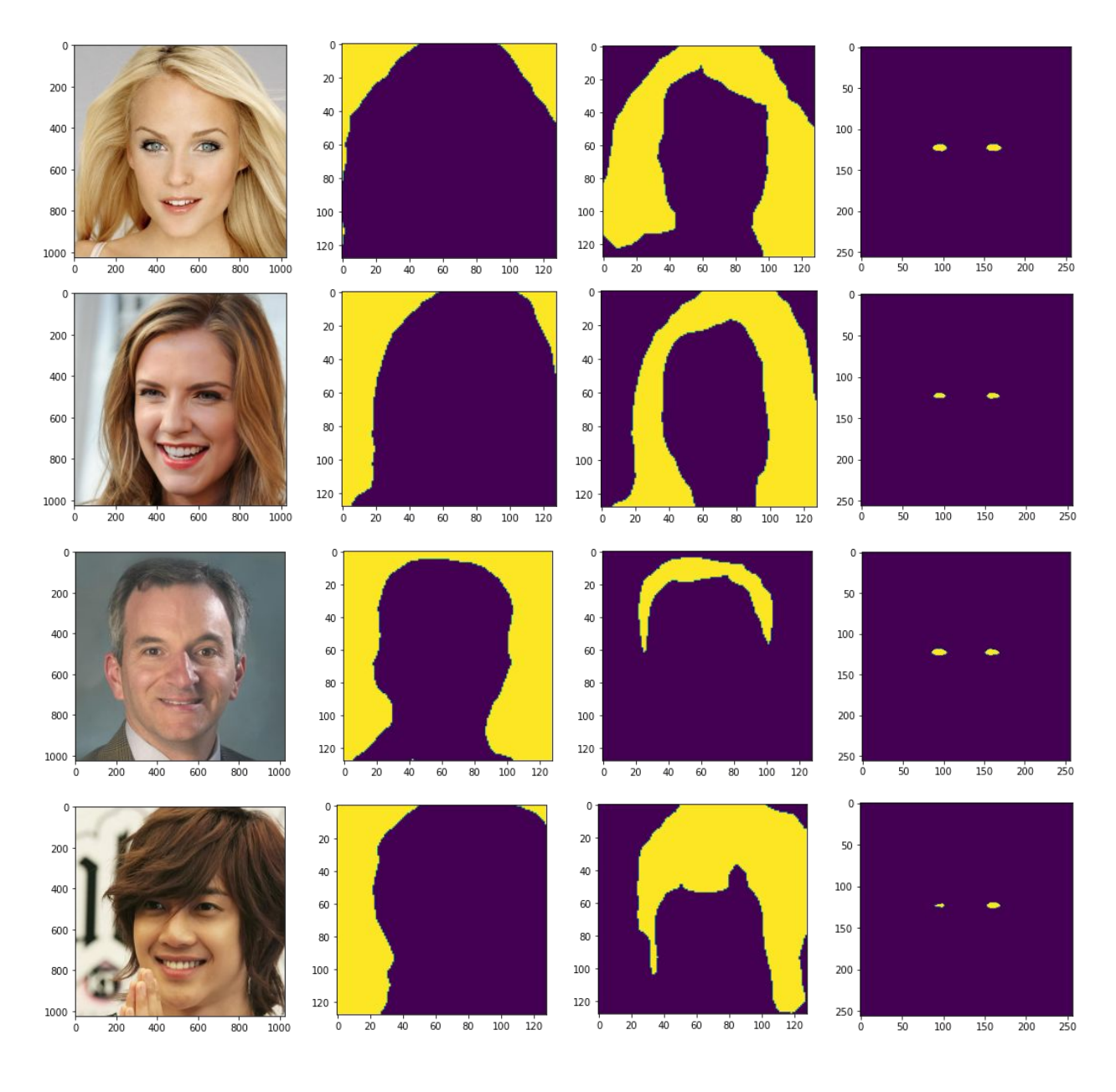}
\end{center}
   \caption{Segmentation results on test set of CelebAMask-HQ~\cite{CelebAMask-HQ} delivered by our segmentation model on semantic classes with a increasing complexity: background, hair, eyes.}
\label{fig:real_segmentation}
\end{figure*}


\section{Method}

Our approach is based on StyleGAN2~\cite{karras2020analyzing} a very powerful generative model capable of generating realistic samples of objects of interest once it has been trained on a related dataset (Section \ref{sec:stylegan}). In our work we either use a pretrained models in case of human faces, cats, dogs and cartoons or train a model from scratch on a desired dataset in case of eye segmentation problem~\cite{palmero2020openeds2020}.

We show that by clustering in the feature space of trained StyleGAN2 model, we can discover meaningful semantic regions that are consistent across different samples of the model (Section \ref{sec:clustering}).

To discover rare semantic classes that were not found during clustering since they appear relatively rarely in samples of StyleGAN2, we apply recently introduced approach for image manipulation~\cite{shen2020interfacegan,patashnik2021styleclip} using text prompts with CLIP~\cite{radford2021learning}. By manipulating latent vectors of sampled images, we make a desired semantic region to appear in almost every sample, which allows it to appear as one of the clusters (Section \ref{sec:manipulation}).

To conclude the process and attribute each semantic region with a prompt defined in a natural language, we modify the original CLIP~\cite{radford2021learning} image encoder to accept image regions instead of a whole image. Then we encode each of the discovered semantic regions and classify them (Section \ref{sec:cluster_classification}).

Once a StyleGAN2 model is modified to deliver samples along with segmentation masks, we apply a simple knowledge distillation procedure by training a segmentation network on synthetic images and segmentation masks and show that it generalizes to real images (Section \ref{sec:distillation}).

\subsection{StyleGAN2}
\label{sec:stylegan}

StyleGAN2 is used in our method as a generative backbone as it is able to generate images with high quality, gives us access to feature maps of generated images that we use for clustering and allows generated images to be manipulated which we use to discover additional semantic classes. Sampling is performed by drawing from a normal distribution $z \in Z$. A separate mapping network maps $z$ to an intermediate latent vector $w \in W$. $w$ is then transformed into $k$ vectors that are used by different layers of generator network of StyleGAN2 (see Fig.~\ref{fig:stylegan_no_manipulation}). StyleGAN2 gradually generates feature maps of higher resolution as they get upsampled and processed by successive layers of the generative model.

\subsection{Clustering}
\label{sec:clustering}

During clustering, we generate $N$ samples and save the generated images as well as corresponding feature maps from a certain layers of StyleGAN. We mostly use feature maps from the seventh or ninth layer of the generative model. Using earlier layers of StyleGan results in more coarse but more semantically meaningful clusters. Later layers result in more fine clusters with less semantic meaning. We also performed experiments with concatenated features from different layers but did not notice any improvement. Feature maps have dimensionality of either $N \times 64 \times 64 \times 512$ or $N \times 128 \times 128 \times 256$ where the first number represents a number of samples the second and third numbers represent the spatial resolution and the last number represents the dimensionality of the feature space. We flatten the acquired feature maps and end up with $N * 64 * 64$ feature maps of size $512$ or with $N * 128 * 128$ feature maps of size $256$. After that we perform clustering by minimizing the following loss function:

$$
J = \sum_{n=1}^{\bar{N}}\sum_{k=1}^{K}r_{nk} \lVert \pmb{x}_n - \pmb{\mu}_k \rVert^2
$$

Where $\pmb{x}_n$ represents a single data point of size $512$ or $256$, $r_{nk} \in {0, 1}$ is a binary indicator variable, where $k= 1, \ldots, K$ describes which of the $K$ clusters the data point $\pmb{x}_n$ is assigned to. $\bar{N}$ is equal to $N * 128 * 128$ or $N * 64 * 64$ depending on the layer of choice.  We search for set of values $\{r_{nk}\}$ and $\{\pmb{\mu}_k\}$ that minimize $J$. To solve this problem we use K-means clustering algorithm~\cite{lloyd1982least}. Discovered classes can be seen in Fig.~\ref{fig:parts}. After the clusters are discovered we notice that new samples have the same semantic regions if we assign their features to the closest cluster. This gives us a way to generate image and corresponding segmentation masks at the same time with minimal additional computational overhead.

\subsection{Class Discovery with Image Manipulation}
\label{sec:manipulation}

\begin{figure}[t]
\begin{center}
\includegraphics[width=1.0\linewidth]{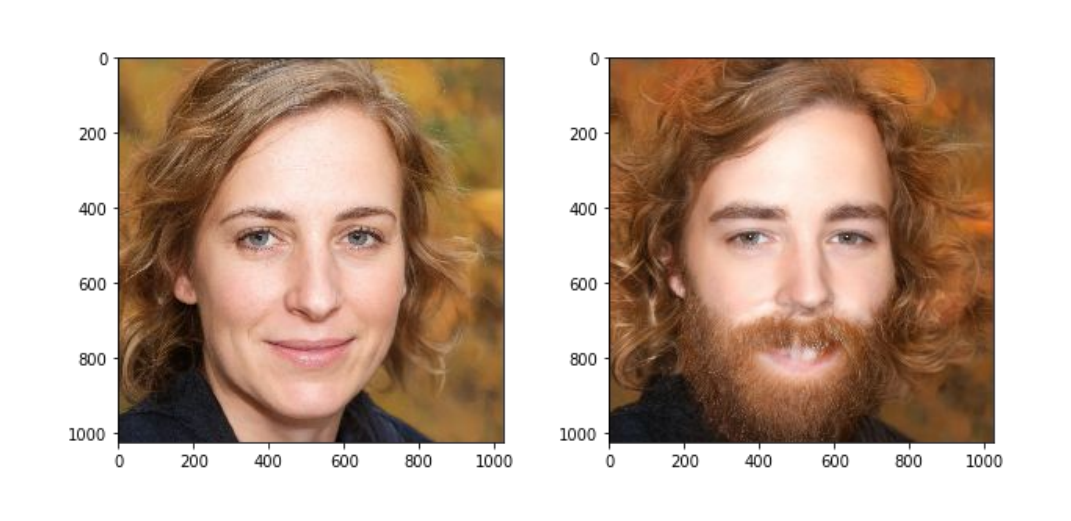}
\end{center}
   \caption{Example of latent code manipulation of generated image along the direction of beard attribute. }
\label{fig:long}
\label{fig:beard_manipulation}
\end{figure}

Some classes that are rarely present in samples of StyleGAN like hat, beard or glasses are not always discovered by our method as is. In order to make desired semantic classes more present in samples that we use for clustering, we update a latent code of every image so that it contains a desired attribute. Let $w \in W$ denote a latent code of a generated sample, and F(w) the corresponding generated image (where F is a generator network of StyleGAN). We employ a recently introduced method~\cite{patashnik2021styleclip,shen2020interfacegan}: we collect a number of latent vectors and corresponding generated images and classify generated images with CLIP~\cite{radford2021learning} given a natural language prompt like ``a person with a beard''. Given a set of pairs $\{(w_i, b_i)\}$, where $w$ is a latent vector and $b$ is a binary variable indicating if the generated images contains desired attribute, we learn a manipulation direction $G$, such
that $F(w + \alpha G)$ yields an image where that attribute is introduced or amplified. The manipulation strength is controlled by $\alpha$. The method depiction can be seen in Fig.~\ref{fig:method_latent}. An example of latent code manipulation along the beard direction can be seen in Fig.~\ref{fig:beard_manipulation} and discovered corresponding semantic regions are depicted in Fig.~\ref{fig:clip_results}.

\subsection{Cluster Classification}
\label{sec:cluster_classification}

Once all the desired clusters are discovered, we need to assign a semantic class to each of them. In order to do so, we use pretrained text encoder and image encoder of the CLIP model~\cite{radford2021learning}. We use text encoder as is and encode desired classes given corresponding text prompts like ``hair``, ``forehead`` etc (see Fig.~\ref{fig:dilated_clip}). As for the image encoder we use a model based on residual neural network~\cite{radford2021learning, he2016deep}. Our choice is motivated by the fact that this model can be modified to deliver embedding with bigger spatial resolution by removing downsampling layers and adding dilation factors to certain convolutional layers~\cite{yu2017dilated, chen2018deeplab}. We observed that bigger spatial resolution gives better results for smaller clusters. Once we get embeddings for each pixel of every image, we average them per each cluster over many images. Averaging over many images also gives better results according to our experiments. Finally, we compute the dot products between embedding of each cluster and each text prompt: we assign the cluster to a class that resulted in a biggest dot product value. This way we use original clusters as region proposals that we classify.

\subsection{Knowledge Distillation}
\label{sec:distillation}

Once we are able to generate synthetic images and corresponding annotations, we need a way to get the same segmentation masks for real images. Inspired by recent work in knowledge distillation for image manipulation with StyleGAN~\cite{viazovetskyi2020stylegan2} we generate a synthetic dataset with images and corresponding segmentation masks and train a simple segmentation model on that dataset. We apply a segmentation method based on resnet-18 and dilated convolutions~\cite{yu2017dilated} and show that trained model also generalizes to real images as it can be seen in Fig.~\ref{fig:real_segmentation}

\section{Datasets and Results}

\begin{figure}[t]
\begin{center}
\includegraphics[width=1.0\linewidth]{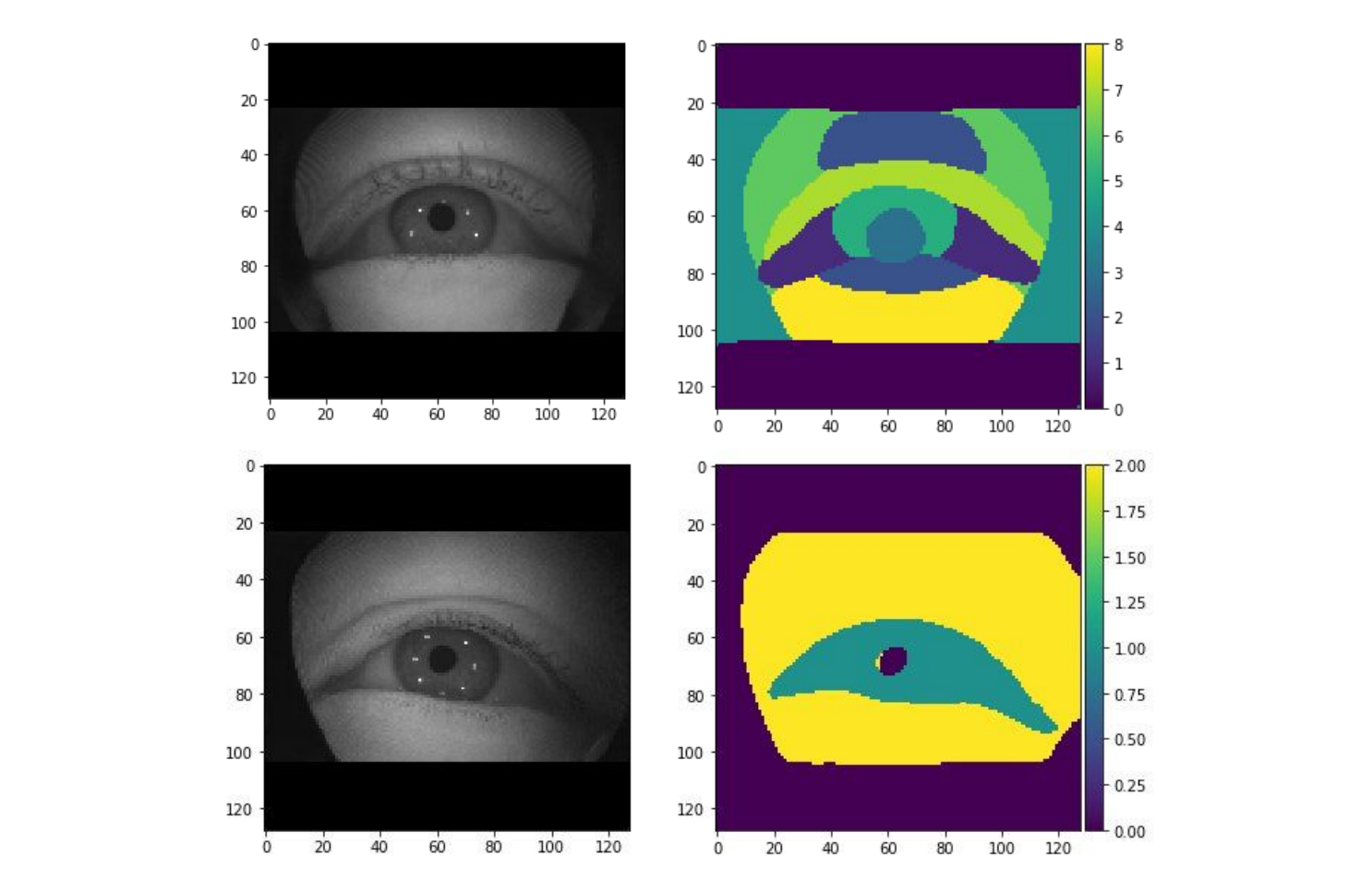}
\end{center}
   \caption{Example of semantic classes discovered by clustering features from different layers for OpenEDS eye segmentation dataset. As it can be seen, overall iris, pupil and sclera were discovered which coincides with semantic classes of the original dataset. }
\label{fig:long}
\label{fig:openeds}
\end{figure}

For faces, we evaluate our model on CelebA-Mask8 dataset~\cite{liu2015deep}, which contains 8 part categories and compare our method to state-of-the-art semi-supervised methods~\cite{zhang2021datasetgan, li2021semantic}. As can be seen in the Table~\ref{tbl:celeb8}  our method achieves state-of-the-art results. In order to evaluate our method on face images that contain a bigger set of labels we test our method on CelebAMask-HQ dataset~\cite{CelebAMask-HQ} segmentation face parsing dataset that contains $19$ semantic classes. To simplify the dataset we merge left and right eye classes into one and do similar simplification for ears. We also eliminate earring and necklace attributes as StyleGAN struggles to generate these attributes realistically which results in poor performance. The results compared to fully supervised method can be found in Table~\ref{tbl:celeb19}. As it can be seen our method is performing worse than the fully supervised method but delivers overall good results. We also collect a small dataset where we label beard in order to test the accuracy of our method on this discovered semantic class. The results compared to fully supervised method can be found in Table~\ref{tbl:beard}. To test if our method is able to generalize to completely different domains, we use our method to perform eye segmentation on OpenEDS~\cite{palmero2020openeds2020} dataset.
We train StyleGAN2 on images of the dataset from scratch and apply our algorithm to the trained model. As it can be seen, overall iris, pupil and sclera were discovered which coincides with semantic classes of the original dataset (see Fig.~\ref{fig:openeds}). Moreover, 
our algorithm achieves a segmentation performance which is very close to the fully supervised method in this case, since the semantic classes are simpler (see Table~\ref{tbl:openeds}).

\section{Conclusion}

We propose a powerful method for unsupervised learning that allows to discover meaningful and consistent semantic classes which mostly coincide with classes defined and labeled by human. Our method is successful in cases where it might be hard for human to define and consistently label semantic classes. We propose a way to derive semantic class annotations by defining a text prompt which allowed us to discover classes like beard that currently has no public annotations. We show that training a segmentation model on our generated synthetic images along with segmentation masks generalizes to real images and shows competitive results with fully supervised methods. Our method achieves state-of-the-art results as it performs better than recently introduced semi-supervised models.

\begin{table*}[t!]
\begin{center}
\begin{tabular}{|l|c|c|c|c|c|c|c|c|}
\hline
Method & DatasetGAN~\cite{zhang2021datasetgan}  &   semanticGAN~\cite{li2021semantic} & Ours\\
\hline 
Number of Manually Annotated Images Used & 16               & 30         & \textbf{0} \\ 
\hline
Mean IOU  & 70.01  & 69.02  & \textbf{73.1}\\ 
\hline

\end{tabular}\\

\caption{ Results of our algorithm on CelebA-Mask8 Dataset~\cite{liu2015deep} in comparison to semi-supervised methods. As it can be seen our method outperforms them and does not need any manual annotations.  } 
\label{tbl:celeb8}
\end{center}
\end{table*}

\begin{table*}[t!]
\begin{center}
\begin{tabular}{|l|c|c|c|c|c|c|c|c|}
\hline
Method & DRN~\cite{yu2017dilated} & Ours\\
\hline 
Number of Manually Annotated Images Used & All & \textbf{0} \\ 
\hline
Mean IOU  & \textbf{70.5} & 62.5 \\ 
\hline

\end{tabular}\\

\caption{ Results of our algorithm on CelebAMask-HQ~\cite{CelebAMask-HQ} in comparison to fully supervised method. The dataset has additional face semantic classes compared to CelebA-Mask8 Dataset~\cite{liu2015deep}.} 
\label{tbl:celeb19}
\end{center}
\end{table*}

\begin{table*}[t!]
\begin{center}
\begin{tabular}{|l|c|c|c|c|c|c|c|c|}
\hline
Method & SegNet~\cite{palmero2020openeds2020} & Ours\\
\hline 
Number of Manually Annotated Images Used & All & \textbf{0} \\ 
\hline
Mean IOU & \textbf{84.1} & 82.39 \\ 
\hline

\end{tabular}\\

\caption{ Results of our algorithm on OpenEDS 2020 eye segmentation dataset~\cite{palmero2020openeds2020} in comparison to fully supervised method. The dataset has $4$ semantic classes. As it can be seen, our method has a very similar performance compared to fully supervised method as semantic classes are simpler in this dataset.} 
\label{tbl:openeds}
\end{center}
\end{table*}

\begin{table*}[t!]
\begin{center}
\begin{tabular}{|l|c|c|c|c|c|c|c|c|}
\hline
Method & DRN~\cite{palmero2020openeds2020} & Ours\\
\hline 
Number of Manually Annotated Images Used & All & \textbf{0} \\ 
\hline
Mean IOU & \textbf{88.8} & 84.29 \\ 
\hline

\end{tabular}\\

\caption{ Results of our algorithm on our custom beard segmentation dataset compared to fully supervised approach.} 
\label{tbl:beard}
\end{center}
\end{table*}

{
    \small
    \bibliographystyle{ieee_fullname}
    \bibliography{macros,main}

\begin{thebibliography}{10}\itemsep=-1pt

\bibitem{pony}
My little pony dataset.
\newblock \url{https://thisponydoesnotexist.net/}.
\newblock Accessed: 2019-09-30.

\bibitem{chen2018deeplab}
Liang-Chieh Chen, George Papandreou, Iasonas Kokkinos, Kevin Murphy, and Alan~L
  Yuille.
\newblock Deeplab: Semantic image segmentation with deep convolutional nets,
  atrous convolution, and fully connected crfs.
\newblock {\em IEEE transactions on pattern analysis and machine intelligence},
  40(4):834--848, 2018.

\bibitem{choi2020stargan}
Yunjey Choi, Youngjung Uh, Jaejun Yoo, and Jung-Woo Ha.
\newblock Stargan v2: Diverse image synthesis for multiple domains.
\newblock In {\em Proceedings of the IEEE/CVF Conference on Computer Vision and
  Pattern Recognition}, pages 8188--8197, 2020.

\bibitem{cordts2016cityscapes}
Marius Cordts, Mohamed Omran, Sebastian Ramos, Timo Rehfeld, Markus Enzweiler,
  Rodrigo Benenson, Uwe Franke, Stefan Roth, and Bernt Schiele.
\newblock The cityscapes dataset for semantic urban scene understanding.
\newblock In {\em Proceedings of the IEEE conference on computer vision and
  pattern recognition}, pages 3213--3223, 2016.

\bibitem{he2016deep}
Kaiming He, Xiangyu Zhang, Shaoqing Ren, and Jian Sun.
\newblock Deep residual learning for image recognition.
\newblock In {\em Proceedings of the IEEE conference on computer vision and
  pattern recognition}, pages 770--778, 2016.

\bibitem{hu2018squeeze}
Jie Hu, Li Shen, and Gang Sun.
\newblock Squeeze-and-excitation networks.
\newblock In {\em Proceedings of the IEEE conference on computer vision and
  pattern recognition}, pages 7132--7141, 2018.

\bibitem{karras2018style}
Tero Karras, Samuli Laine, and Timo Aila.
\newblock A style-based generator architecture for generative adversarial
  networks.
\newblock {\em arXiv preprint arXiv:1812.04948}, 2018.

\bibitem{karras2020analyzing}
Tero Karras, Samuli Laine, Miika Aittala, Janne Hellsten, Jaakko Lehtinen, and
  Timo Aila.
\newblock Analyzing and improving the image quality of stylegan.
\newblock In {\em Proceedings of the IEEE/CVF Conference on Computer Vision and
  Pattern Recognition}, pages 8110--8119, 2020.

\bibitem{CelebAMask-HQ}
Cheng-Han Lee, Ziwei Liu, Lingyun Wu, and Ping Luo.
\newblock Maskgan: Towards diverse and interactive facial image manipulation.
\newblock In {\em IEEE Conference on Computer Vision and Pattern Recognition
  (CVPR)}, 2020.

\bibitem{li2021semantic}
Daiqing Li, Junlin Yang, Karsten Kreis, Antonio Torralba, and Sanja Fidler.
\newblock Semantic segmentation with generative models: Semi-supervised
  learning and strong out-of-domain generalization.
\newblock In {\em Proceedings of the IEEE/CVF Conference on Computer Vision and
  Pattern Recognition}, pages 8300--8311, 2021.

\bibitem{liu2015deep}
Ziwei Liu, Ping Luo, Xiaogang Wang, and Xiaoou Tang.
\newblock Deep learning face attributes in the wild.
\newblock In {\em Proceedings of the IEEE international conference on computer
  vision}, pages 3730--3738, 2015.

\bibitem{lloyd1982least}
Stuart Lloyd.
\newblock Least squares quantization in pcm.
\newblock {\em IEEE transactions on information theory}, 28(2):129--137, 1982.

\bibitem{palmero2020openeds2020}
Cristina Palmero, Abhishek Sharma, Karsten Behrendt, Kapil Krishnakumar, Oleg~V
  Komogortsev, and Sachin~S Talathi.
\newblock Openeds2020: open eyes dataset.
\newblock {\em arXiv preprint arXiv:2005.03876}, 2020.

\bibitem{patashnik2021styleclip}
Or Patashnik, Zongze Wu, Eli Shechtman, Daniel Cohen-Or, and Dani Lischinski.
\newblock Styleclip: Text-driven manipulation of stylegan imagery.
\newblock {\em arXiv preprint arXiv:2103.17249}, 2021.

\bibitem{radford2021learning}
Alec Radford, Jong~Wook Kim, Chris Hallacy, Aditya Ramesh, Gabriel Goh,
  Sandhini Agarwal, Girish Sastry, Amanda Askell, Pamela Mishkin, Jack Clark,
  et~al.
\newblock Learning transferable visual models from natural language
  supervision.
\newblock {\em arXiv preprint arXiv:2103.00020}, 2021.

\bibitem{shen2020interfacegan}
Yujun Shen, Ceyuan Yang, Xiaoou Tang, and Bolei Zhou.
\newblock Interfacegan: Interpreting the disentangled face representation
  learned by gans.
\newblock {\em IEEE transactions on pattern analysis and machine intelligence},
  2020.

\bibitem{viazovetskyi2020stylegan2}
Yuri Viazovetskyi, Vladimir Ivashkin, and Evgeny Kashin.
\newblock Stylegan2 distillation for feed-forward image manipulation.
\newblock In {\em European Conference on Computer Vision}, pages 170--186.
  Springer, 2020.

\bibitem{xie2017aggregated}
Saining Xie, Ross Girshick, Piotr Doll{\'a}r, Zhuowen Tu, and Kaiming He.
\newblock Aggregated residual transformations for deep neural networks.
\newblock In {\em Proceedings of the IEEE conference on computer vision and
  pattern recognition}, pages 1492--1500, 2017.

\bibitem{yu2017dilated}
Fisher Yu, Vladlen Koltun, and Thomas Funkhouser.
\newblock Dilated residual networks.
\newblock In {\em Proceedings of the IEEE conference on computer vision and
  pattern recognition}, pages 472--480, 2017.

\bibitem{zhang2021datasetgan}
Yuxuan Zhang, Huan Ling, Jun Gao, Kangxue Yin, Jean-Francois Lafleche, Adela
  Barriuso, Antonio Torralba, and Sanja Fidler.
\newblock Datasetgan: Efficient labeled data factory with minimal human effort.
\newblock In {\em Proceedings of the IEEE/CVF Conference on Computer Vision and
  Pattern Recognition}, pages 10145--10155, 2021.

\bibitem{zhao2017pyramid}
Hengshuang Zhao, Jianping Shi, Xiaojuan Qi, Xiaogang Wang, and Jiaya Jia.
\newblock Pyramid scene parsing network.
\newblock In {\em Proceedings of the IEEE conference on computer vision and
  pattern recognition}, pages 2881--2890, 2017.

\end{thebibliography}
}



\end{document}